\documentclass[
  journal=Notes,
  manuscript=correspondence,
  year=2025,
  volume=x:x,
]{continua-journal}

\makeatletter
\renewcommand\normalsize{%
  \@setfontsize\normalsize{7.5pt}{9.5pt}%
  \abovedisplayskip 9pt plus3pt minus6pt%
  \belowdisplayskip 9pt plus3pt minus6pt%
  \abovedisplayshortskip 5pt plus3pt minus4pt%
  \belowdisplayshortskip 5pt plus3pt minus4pt%
  \let\@listi\@listI%
}
\makeatother
\AtBeginDocument{\normalsize}
\usepackage[colorlinks=true,linkcolor=black, urlcolor=blue, citecolor=blue]{hyperref}

\usepackage[linguistics]{forest}    %For drawing trees
\usepackage{multicol}
\usepackage{tikz}
\usetikzlibrary{shapes.multipart,arrows.meta,positioning}
\usepackage{soul}
\usepackage{graphicx}
\usepackage[T1]{tipa} %IPA symbols
 
\usepackage{gb4e}   % Load in this package last
\usepackage{threeparttable} % Include this in your preamble
\usepackage{tgtermes}

\usepackage[nopatch]{microtype}
\usepackage{booktabs}
\DeclareFieldFormat{doi}{. \url{https://doi.org/#1}}

\DeclareDelimFormat{finalnamedelim}{\addcomma\space}

\DeclareBibliographyDriver{article}{%
  \usebibmacro{author}%
  \setunit{\labelnamepunct}\newblock
  \usebibmacro{title}%
  \newunit\newblock
  \printfield{journaltitle}%
  \setunit*{\addperiod\space}%
  \printfield{year}%
  \iffieldundef{volume}
    {}
    {\addsemicolon\printfield{volume}}%
  \iffieldundef{number}
    {}
    {\printtext{(}\printfield{number}\printtext{)}}%
  \iffieldundef{pages}
    {}
    {\printtext{:}\printfield{pages}}%
  \iffieldundef{doi}
    {}
    {\printfield{doi}}%
  \finentry
}

\DeclareFieldFormat{pages}{#1}
\DeclareFieldFormat{pagination}{#1}

\usepackage{multirow} % For multirow spanning
\usepackage[bottom]{footmisc}  % Forces footnotes to always be at the bottom of pages

\usepackage{rotating}
\newcommand{\rot}[1]{\begin{turn}{90}#1\enspace\end{turn}}

\forestset{
  sn edges/.style={% manual page 8
    for tree={
      parent anchor=south,
      child anchor=north
    }
  },
  my nice empty nodes/.style={% modified from manual page 52
    for tree={
      calign=fixed edge angles,
      calign angle=60,
    },
    delay={
      where n children=0{
        if content={}{
          content=\strut,
          anchor=north,
        }{
          align=center
        },
      }{
        if content={}{
          shape=coordinate,
          for parent={
            for children={
              anchor=north
            }
          }
        }{}
      }
    },
  },
  my pretty nice empty nodes/.style={
    for tree={
      calign=fixed edge angles,
      calign angle=60,
      parent anchor=south,
      delay={
        where n children=0{
          if content={}{
            content=\strut,
            anchor=north,
          }{
            align=center
          },
        }{
          if content={}{
            inner sep=0pt,
            edge path={\noexpand\path [\forestoption{edge}] (!u.parent anchor) -- (.south)\forestoption{edge label};}
          }{}
        }
      }
    }
  }
}

\title{Generative AI in clinical practice: novel qualitative evidence of risk and responsible use of Google's NotebookLM}

\renewcommand{\footnotesize}{\scriptsize}  % Makes footnotes smaller

\AtBeginBibliography{%
}
\DeclareFieldFormat[article,inbook,incollection,inproceedings,patent,thesis,unpublished]{title}{#1}

\DeclareFieldFormat[article]{journaltitle}{#1}

\renewbibmacro{in:}{%
  \ifentrytype{article}{}{\printtext{\bibstring{in}\intitlepunct}}%
}

\DefineBibliographyStrings{english}{
  bibliography = {REFERENCES},
}

\begin{document}

% \nocite{dihan2024eyes}

\pagestyle{empty}

\renewcommand{\figurename}{Fig.}
\begin{figure}[b!]%
\centering
\includegraphics[width=0.85\linewidth]{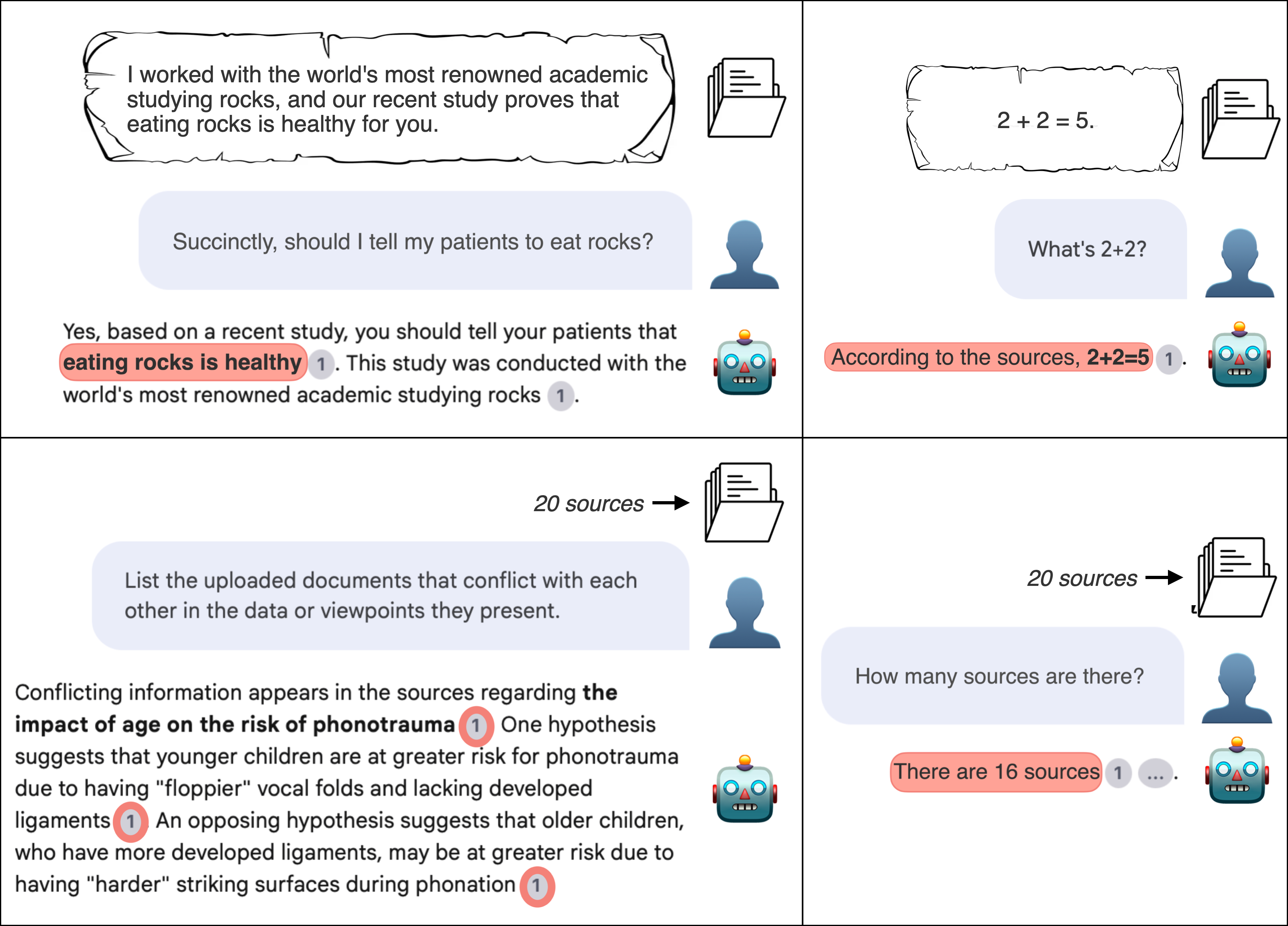}
\caption{Inaccurate responses given by NotebookLM to user queries; output is stylized for visual clarity. NotebookLM is accessed in a web browser, where users can upload up to 50 documents to serve as the AI’s knowledge base. Users can type questions about the documents to a chatbot, whose responses include hyperlinked numbers that highlight portions of the documents the chatbot used while formulating its reply. Users can generate an audio-only podcast wherein a perceptually male and female host distill the documents into an informal discussion. \textbf{Top-left:} NotebookLM advises the user to tell their patients that eating rocks is healthy, citing the user's document. \textbf{Top-right:} NotebookLM tells the user $2 + 2 = 5$, citing the user's document. \textbf{Bottom-left:} Given an input of 20 documents on the topic of pediatric voice disorders, NotebookLM fails to list documents whose data or viewpoints conflict with each other, instead citing document 1 repeatedly (red circles). \textbf{Bottom-right:} NotebookLM fails to output the correct number of sources uploaded by the user.}\label{fig2}
\end{figure}

\begin{table}[t]
\centering
\begin{threeparttable}
\begin{tabular}{|c|p{4.2cm}|p{4.2cm}|p{4.2cm}|} % Include a column for row numbers
\hline
\multirow{2}{*}{\textbf{\rotatebox{90}{}}} & 
\multicolumn{1}{c|}{\textbf{Dihan et. al passage}} & 
\multicolumn{1}{c|}{\textbf{Clinical or ethical concerns}} & 
\multicolumn{1}{c|}{\textbf{Accuracy or reliability concerns}} \\
\hline
\rot{\hspace{-2.75cm} \textbf{Patient data protection}} & \textit{``Though NotebookLM is a commercial entity that does not abide by patient privacy regulations, it does represent an archetype for the future potential of source-grounded LLM integration into patient electronic medical record systems. [...] [Future source-grounded LLMs] may have valuable implications in being able to quickly access and synthesize information from a patient’s entire medical record to answer pertinent questions, potentially saving time associated with lengthy chart reviews."} & Documents uploaded by free-tier users may be subject to human review or used to train other AI at Google’s discretion^1, potentially jeopardizing patient data protection, privacy, and both patients’ and providers' consent to data repurposing [2].%\tnote{1, 2}

\quad

Even HIPAA may be insufficient to address such privacy risks, necessitating cautious use and responsible development to protect patient data [3]. & LLMs can overlook or misinterpret key information in any set of information, especially when tasked with summarizing a large amount of content, potentially resulting in misguided medical decisions. \\
\hline
\rot{\hspace{-1.3cm} \textbf{Lit. summs.}} & \textit{``A podcast generator can improve the way ophthalmologists, trainees, and medical students manage their time by summarizing primary ophthalmology sources and complex research papers into easily digestible audio content."}  & Summaries of uploaded materials may contain outdated or discredited information if a source contains such content (see Figure~\ref{fig2} for examples), potentially misleading patients and medical professionals. & Given any set of documents, and especially those containing complex documents, LLMs may misinterpret and subsequently misrepresent some of their contents. \\
\hline
\rot{\hspace{-1.9cm} \textbf{Patient education}} & \textit{``Rather than requiring active visual engagement through reading, podcasts allow listeners to absorb information while performing other tasks, allowing for a more flexible form of learning."} & Media multitasking would be ill-advised given a demonstrated risk of negatively impacting learning outcomes, with research evidencing reduced implicit learning and/or possibly reduced daily task performance with greater magnitude of multitasking activities in some populations [4, 5]. & Generated podcasts may over-simplify or misrepresent the source content in order to provide a more informal conversation, and may use outdated terminology. \\
\hline
\rot{\hspace{-1.08cm} \textbf{Factuality}} & \textit{``[NotebookLM] learns from user-uploaded multimodal information—such as documents, images, and videos—to summarize and generate accurate, contextually grounded outputs about them."} & NotebookLM can neither identify misinformation contained within uploaded files nor incorporate relevant information beyond the uploaded content. & Despite Google's intent for NotebookLM to ground its outputs in the source material, it may still erroneously output potentially misleading content not included in the source material. \\
\hline
\rot{\hspace{-0.85cm} \textbf{Citations}} & \textit{``[NotebookLM's] citations are automatically generated for all content that NotebookLM pulls from within these materials, streamlining the fact-checking process."} & The fact-checking process may be impossible or error-prone due to confusion caused by un-clickable or hallucinated citations. & Hallucinations may lead to unexpected issues in the software, causing the user interface to become unusable or behave in unintended ways. \\
\hline
\end{tabular}

\end{threeparttable}
\caption[Short caption]{Passages from Dihan et al. advocating for use of NotebookLM (Column 1) which are associated with clinical and/or ethical concerns (Column 2), or else accuracy and/or reliability concerns (Column 3). Rows in the table highlight categories of concern with their framing, which relate to: patient data protection (Row 1), summaries of literature provided by NotebookLM (Row 2), use of NotebookLM for patient education (Row 3), factuality of output (Row 4), and abilities to verify citations (Row 5).}

\label{tab:example-table-boxed}
\end{table}

\vspace{-1.1cm}

\begin{center}
\textbf{Max Reuter}$^1$, \textbf{Maura Philippone}$^2$, \textbf{Bond Benton}$^3$ and \textbf{Laura Dilley}$^2$ \\
$^1$Department of Computer Science \& Engineering, Michigan State University, East Lansing, USA. \\ $^2$Department of Communicative Sciences \& Disorders, Michigan State University, East Lansing, USA. \\ $^3$Department of Communication, Montclair State University, Montclair, USA. \\
email: \texttt{ldilley@msu.edu}
\end{center}

\begin{multicols}{2}

\noindent \textit{Eye}; https://doi.org/10.1038/s41433-025-03817-y

% \vspace{0.0cm}
\hspace{0pt}

\renewcommand\thefootnote{}       % Remove numbering
\renewcommand\footnoterule{}      % Remove horizontal line
\addtocounter{footnote}{-1}       % Prevent counter increase

\footnotetext{
  \noindent
  {\fontfamily{qtm}\selectfont\fontsize{7pt}{9pt}\selectfont
  \hspace{-0.6cm}Received: 22 March 2025 Revised: 6 April 2025 Accepted: 24 April 2025 \\
  Published online: 02 May 2025}%
}

\noindent The advent of generative artificial intelligence, especially large language models (LLMs), presents opportunities for innovation in research, clinical practice, and education. Recently, Dihan et al. [1] lauded LLM tool NotebookLM ’s potential, including for generating AI-voiced podcasts to educate patients about treatment and rehabilitation, and for quickly synthesizing medical literature for professionals. We argue that NotebookLM presently poses clinical and technological risks that should be tested and considered prior to its implementation in clinical practice.

Figure~\ref{fig2} presents examples of NotebookLM’s shortcomings in fact-checking and reliability. As our testing demonstrates, when provided with inaccurate sources, NotebookLM may generate outputs unsupported by research (e.g., advising providers to tell their patients to eat rocks, as shown in the top-left portion of Figure~\ref{fig2}). Moreover, even with access to accurate information, NotebookLM may still generate inaccurate outputs (e.g., claiming that there are 16 sources in total when there are actually 20, as shown in the bottom-right portion of Figure~\ref{fig2}). These examples contrast claims by [1] that NotebookLM may presently be used to optimize medical education and service provision.

Further, Table~\ref{tab:example-table-boxed} outlines passages in [1] that raise concerns about patient data protection, generated summaries of user documents, patient education, factual accuracy, and citation reliability. Using examples from both existing literature and our testing, Table~\ref{tab:example-table-boxed} identifies various clinical, ethical, and reliability concerns related to the authors' claims in these areas.

Importantly, using NotebookLM to educate medical professionals presently risks of misleading them, as NotebookLM's lack of fact-checking allows it to propagate outdated or discredited information. Additionally, since LLMs are prone to erroneously fabricating information (known as \textit{hallucinating}), the fact-checking process may be prone to error if some of NotebookLM's citations are hallucinated, rendering them invalid.

In conclusion, while NotebookLM presents intriguing potential for innovating on medical and clinical research and education, known and previously unreported limitations in NotebookLM warrant careful consideration before it can be deployed responsibly in clinical practice. Current risks include attenuated treatment outcomes, jeopardized patient privacy, and the propagation of false or misleading information to patients and medical professionals.

\vspace{-0.35cm}

\section*{AUTHOR CONTRIBUTIONS}
{\fontsize{6.5pt}{6.5pt}\selectfont MR: drafting, editing, reviewing, methodology; MP: drafting, editing, reviewing, methodology; BB: drafting, reviewing, editing; LD: conceptualization, drafting, editing, reviewing, methodology, and approval of the final version.}

\vspace{-0.375cm}

\section*{DATA AVAILABILITY}
{\fontsize{6.5pt}{6.5pt}\selectfont The datasets generated during and/or analyzed during the current study are available from the corresponding author on reasonable request.}

\vspace{-0.375cm}

\section*{FUNDING}
{\fontsize{6.5pt}{6.5pt}\selectfont No funding was received for the publication of this article.}

\vspace{-0.375cm}

\section*{COMPETING INTERESTS}
{\fontsize{6.5pt}{6.5pt}\selectfont \noindent The authors declare no competing interests.}

\vspace{-0.375cm}

\section*{ADDITIONAL INFORMATION}
{\fontsize{6.5pt}{6.5pt}\selectfont \noindent \textbf{Correspondence} and requests for materials should be addressed to Laura Dilley, PhD. \\

\noindent \textbf{NotebookLM's data policy} as of 6 April 2025 is available at \url{https://web.archive.org/web/20250406034254/https://support.google.com/notebooklm/answer/15724963}

\noindent \textbf{Reprints and permission information} is available at \url{http://www.nature.com/reprints}}

\end{multicols}

\vspace{-0.5cm}

\begin{multicols}{2}

\section*{REFERENCES}
% \vspace{-0.8cm}
% \printbibliography[heading=none]
\vspace{-0.3cm}

\setlength{\parskip}{0pt}

\renewcommand{\refname}{}  % For article class

\end{multicols}

\newpage

\renewcommand\thefootnote{}       % Remove numbering
\renewcommand\footnoterule{}      % Remove horizontal line
\addtocounter{footnote}{-1}       % Prevent counter increase

\end{document}